\documentclass[letterpaper, 10 pt, conference]{ieeeconf}  % Comment this line out if you need a4paper

\IEEEoverridecommandlockouts                              % This command is only needed if 
\usepackage{graphics} % for pdf, bitmapped graphics files
\usepackage{epsfig} % for postscript graphics files
\usepackage{cite}
\usepackage{url}
\usepackage{xcolor}
\usepackage{gensymb}
\usepackage{amsmath}
\usepackage{float}

\title{\LARGE \bf
Influence of Gripper Design on Human Demonstration Quality for Robot Learning
}

\author{Gina L. Georgadarellis, Natalija Beslic, Seonhun Lee, \\
Frank C. Sup IV, \IEEEmembership{Member, IEEE}, and Meghan E. Huber, \IEEEmembership{Member, IEEE}
\thanks{This work was supported by the Elaine Marieb Center for Nursing and Engineering Innovation. Gina L. Georgadarellis was also supported by a Robert and Deanna Hagerty Scholarship and the Prof. John R. Dixon Fellowship from the University of Massachusetts Amherst.}
\thanks{All authors are with the Department of Mechanical and Industrial Engineering, University of Massachusetts Amherst, Amherst, MA 01003 USA.}
\thanks{Gina L. Georgadarellis, Seonhun Lee, and Frank C. Sup IV are also with the Elaine Marieb Center for Nursing and Engineering Innovation, University of Massachusetts Amherst, Amherst, MA 01003 USA.}
\thanks{Frank C. Sup IV and Meghan E. Huber are also with the Center for Personalized Health Monitoring, Institute for Applied Life Sciences, University of Massachusetts Amherst, Amherst, MA 01003.}
\thanks{Email: \{ggeorgad, seonhunlee, sup, mehuber\}@umass.edu and \mbox{natalija@beslic.com}}}

\begin{document}

\bstctlcite{IEEEexample:BSTcontrol}

\maketitle
\thispagestyle{empty}
\pagestyle{empty}

%%%%%%%%%%%%%%%%%%%%%%%%%%%%%%%%%%%%%%%%%%%%%%%%%%%%%%%%%%%%%%%%%%%%%%%%%%%%%%%%
\begin{abstract}

Opening sterile medical packaging is routine for healthcare workers but remains challenging for robots. Learning from demonstration enables robots to acquire manipulation skills directly from humans, and handheld gripper tools such as the Universal Manipulation Interface (UMI) offer a pathway for efficient data collection. However, the effectiveness of these tools depends heavily on their usability. We evaluated UMI in demonstrating a bandage opening task, a common manipulation task in hospital settings, by testing three conditions: distributed load grippers, concentrated load grippers, and bare hands. Eight participants performed timed trials, with task performance assessed by success rate, completion time, and damage, alongside perceived workload using the NASA-TLX questionnaire. Concentrated load grippers improved performance relative to distributed load grippers but remained substantially slower and less effective than hands. These results underscore the importance of ergonomic and mechanical refinements in handheld grippers to reduce user burden and improve demonstration quality, especially for applications in healthcare robotics.

\end{abstract}

%%%%%%%%%%%%%%%%%%%%%%%%%%%%%%%%%%%%%%%%%%%%%%%%%%%%%%%%%%%%%%%%%%%%%%%%%%%%%%%%
\section{INTRODUCTION}
\label{section: introduction}

There is a nationwide shortage of health workers, and with a projected gap of 11 million workers by 2030 \cite{WHO2025}, this growing workforce challenge presents a unique opportunity for roboticists and designers to create impactful technologies that enhance efficiency and improve patient care. While there are some concerns over use of robotic technology in healthcare such as potential job loss and a lack of personal touch and empathy, nurses – who comprise the largest group of healthcare workers \cite{NAP25982} and are direct end users of medical devices – have a positive perception of robotic technology \cite{GG2024surveyPaper} and even welcome this technology when it improves workflow \cite{lee2018nurses}. 

Robot learning from human demonstration offers a particularly promising avenue for healthcare robotics because it enables robots to acquire advanced manipulation skills directly from demonstrations by human healthcare workers without requiring programming expertise \cite{ravichandar2020recent}. In fact, recent advances in learning from demonstration have introduced innovative methods for capturing demonstrations that offer more intuitive and efficient transfer of skills to robots. For example, human demonstrators have successfully trained robots in pick-and-place and other manipulation tasks using wearable exoskeleton gloves \cite{Lu2023parallalhandexo}, teleoperation systems such as ALOHA \cite{zhao2023learning}, and handheld gripper tools like “reacher-grabber sticks” \cite{shafiullah2023bringing} and the Universal Manipulation Interface (UMI) \cite{chi2024universal}. These tools allow humans to demonstrate tasks ``in-the-wild" while closely replicating the interaction dynamics and sensing capabilities of the robot’s gripper, thereby enabling demonstrations to transfer more directly into deployable robot control policies.

\begin{figure}[t]
 \centering  \includegraphics[width=\columnwidth]{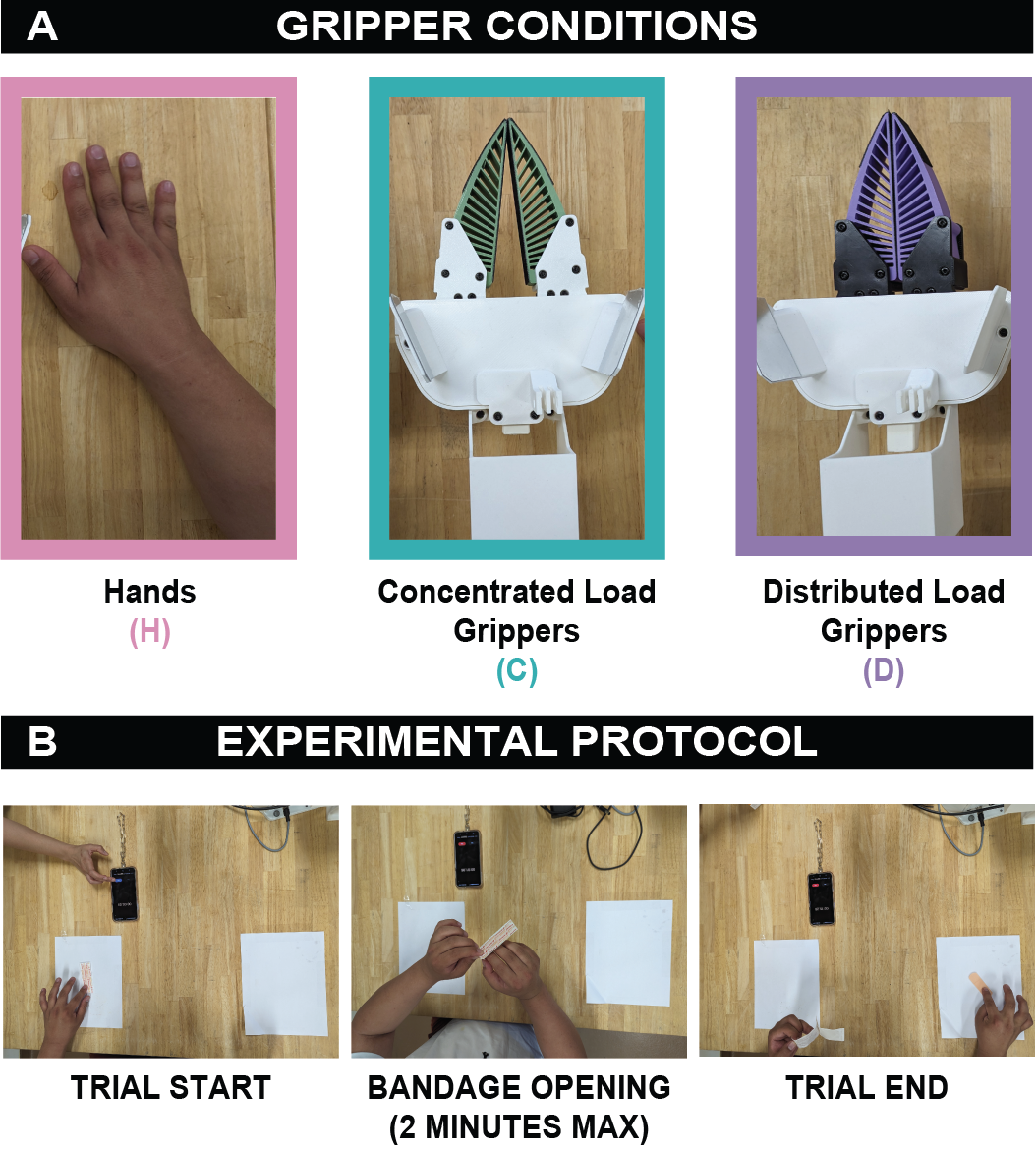}
  \caption{\textbf{A: Gripper Conditions.} Participants bimanually opened bandages with three different gripper conditions.  \textbf{B: Experimental Protocol.} In each trial, participants opened a bandage and removed it from its wrapper.}
  \label{fig: taskDescription}
\end{figure}

While handheld gripper tools show promise for teaching robots manipulation skills, realizing this potential to improve workflow in healthcare settings will require overcoming additional challenges. In particular, healthcare workers must be able to use these tools to perform tasks that go beyond simple pick-and-place, such as removing complex packaging, preparing materials, and handling medical supplies, without the tools impeding their ability to complete the tasks efficiently. This is important not only to encourage healthcare workers to provide demonstrations, but also to ensure that robots learn to perform these tasks efficiently.

The goal of this study was to evaluate the usability of handheld gripper tools for demonstrating healthcare-relevant manipulation tasks to enable robot learning (Figure \ref{fig: taskDescription}). One area where robotic technology can assist in healthcare is the manipulation of commonly used medical supplies. Types of interactions associated with medical supplies can range from autonomous control that involves no human interaction (e.g., restocking the supply closet, kit making, item disposal), nurse only (e.g. package opening and hand-off, fetching and delivering), and patient-facing tasks (e.g. using the item on the patient, wrapping gauze, placing electrodes). Because nurses are interested in preserving personal touch and human-human interaction, this study investigates manipulating medical supplies for non-patient-facing scenarios. Specifically, bandage opening was selected as the demonstration task because it reflects a common clinical need involving sterile packaged objects and exemplifies the manipulation challenges frequently encountered in healthcare. 

As detailed in Section \ref{section: gripper}, our initial pilot testing revealed immediate difficulties in performing this task with existing handheld gripper tools. This is perhaps unsurprising, as such tools are typically designed for pick-and-place and interaction with primarily rigid objects rather than the fine manipulation of thin, flexible materials required in this task. Thus, we also sought to determine whether modifications to the mechanical design of these tools could improve their usability, and in turn, the quality and timeliness of demonstrations. As a result, the contribution of this work is a systematic evaluation of handheld gripper usability in a healthcare-relevant manipulation task, including an analysis of how design modifications affect demonstration performance and perceived workload by the demonstrator.

The remainder of this paper is organized as follows. Section II introduces the handheld gripper tool used in the usability study and motivates the design modifications evaluated. Section III describes the experimental methods of the usability study, and Section IV presents the results. Section V discusses the findings and their implications for healthcare robotics, and Section VI concludes the paper.
\section{Overview of UMI and Its Design Challenges}
\label{section: gripper}

The Universal Manipulation Interface (UMI) stands out among recent advances for its portable in-the-wild data collection capabilities and an integrated policy learning pipeline that enables direct transfer of human demonstrations into deployable robot control policies \cite{chi2024universal}. The handheld gripper tool features two compliant parallel fingers actuated by a single spring-loaded trigger, allowing the human demonstrator to control finger position. A camera mounted on each handheld tool provides visual sensing of both the gripper and the external environment. To facilitate skill transfer, the robot is equipped with a similar compliant parallel-finger gripper, enabling it to replicate the demonstrated actions with comparable interaction dynamics. With the feasibility and promise of handheld gripper tools demonstrated, the next step toward adoption is to identify their current limitations and explore ways to improve their design and effectiveness. 

\subsection{Performance and Usability Limitations}

Despite their promise, the developers of UMI acknowledge several factors that may limit the performance and usability of this and similar systems. First, the demonstrations for robot learning are already time-consuming, and handheld gripper tools can make them take even longer. The current learning framework of UMI, for example, requires at least 200 demonstrations to train a single task in a fixed environment. Although UMI demonstrations are faster than those collected via teleoperation, they still take roughly twice as long as using hands \cite{chi2024universal}. Second, usability challenges further limit performance, as well as user comfort. For instance, each UMI handheld gripper tool weighs about 780 g \cite{chi2024universal}, which can cause fatigue during prolonged use and negatively affect the demonstration.

Recent work has begun exploring improvements to the mechanical design of UMI to enhance the robot learning pipeline \cite{liu2024fastumi}. However, the broader challenge of improving the mechanical design and ergonomics of handheld gripper tools remains relatively underexplored, particularly with respect to demonstration quality and the effort required from a human demonstrator. With a view towards using robot learning from human demonstration in healthcare settings, addressing these challenges is especially vital. Healthcare professionals, such as nurses, already face high cognitive and physical demands, so technology that adds further burden could hinder rather than support adoption.

\begin{figure}[t]
    \centering
    \vspace{1.5mm}
    \includegraphics[width=\linewidth]{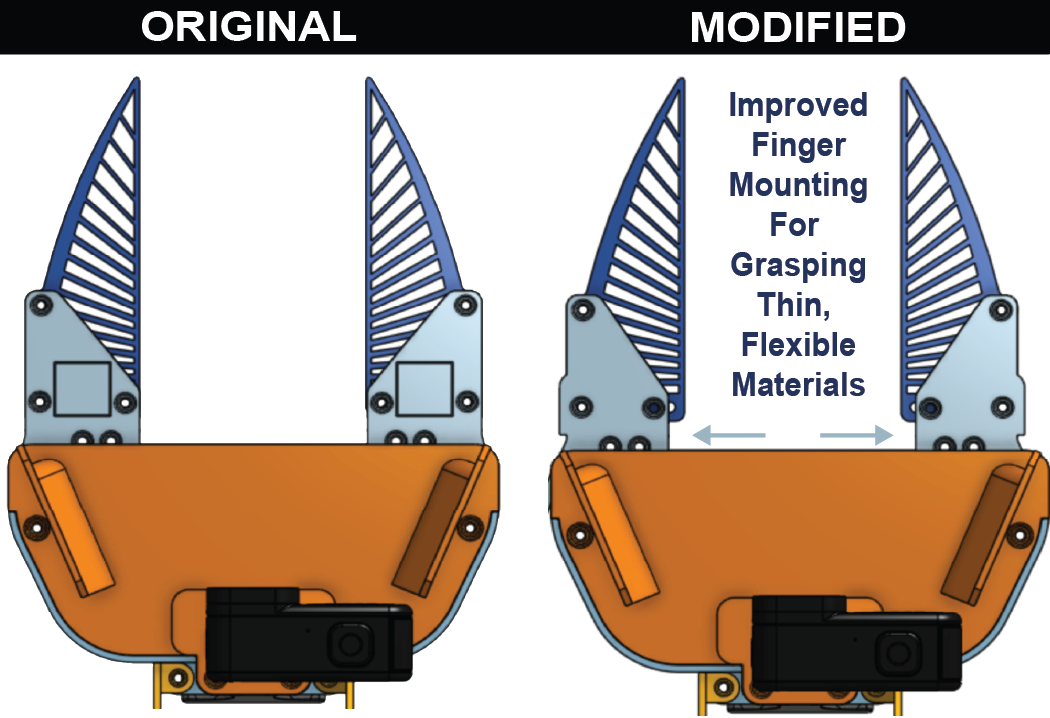}
    \caption{\textbf{Redesigned finger mount of the UMI handheld gripper.} The modification allows the fingers to make direct contact when fully closed, enabling effective grasping.}
    \label{fig:fingerMountModification}
\end{figure}

\subsection{Design Modifications UMI for Bandage Opening}

In our initial pilot testing of UMI in its original hardware configuration, demonstrators were unable to complete the bandage opening task. We identified two main causes. First, the compliant fingers could not grasp the bandage because, when fully closed, the rigid mounting components contacted each other before the fingers did, leaving a gap larger than the packaged bandage. As a result, the gripper could not apply sufficient compressive force. To address this, we redesigned the finger mount so the fingers make direct contact when fully closed, enabling effective grasping (Figure \ref{fig:fingerMountModification}). Second, the compliant nature of the fingers made fine manipulation extremely difficult. While the exact reason remains unclear, it may be due to limitations in either force application or tactile sensing by the demonstrators. A possible explanation, based on anecdotal observations during pilot testing, is that the packaged bandage behaves as a flexible planar object, which prevents the compliant fingers from deforming during grasp. As a result, force could only be applied within a small region near the proximal end of the finger, requiring precise contact placement. This challenge was compounded by the bimanual nature of the task, as the grippers obscured the demonstrator’s view and made accurate placement difficult, preventing bandages from being opened within the allotted time. To address this issue, the finger material was changed from compliant thermoplastic polyurethane (TPU) to rigid polylactic acid (PLA), enabling more evenly distributed force along the contact surface and allowing demonstrators to use the full length of the finger without obscuring their view.

\begin{figure*}
   \centering
   \vspace{1.5mm}
   \includegraphics[width=\textwidth]{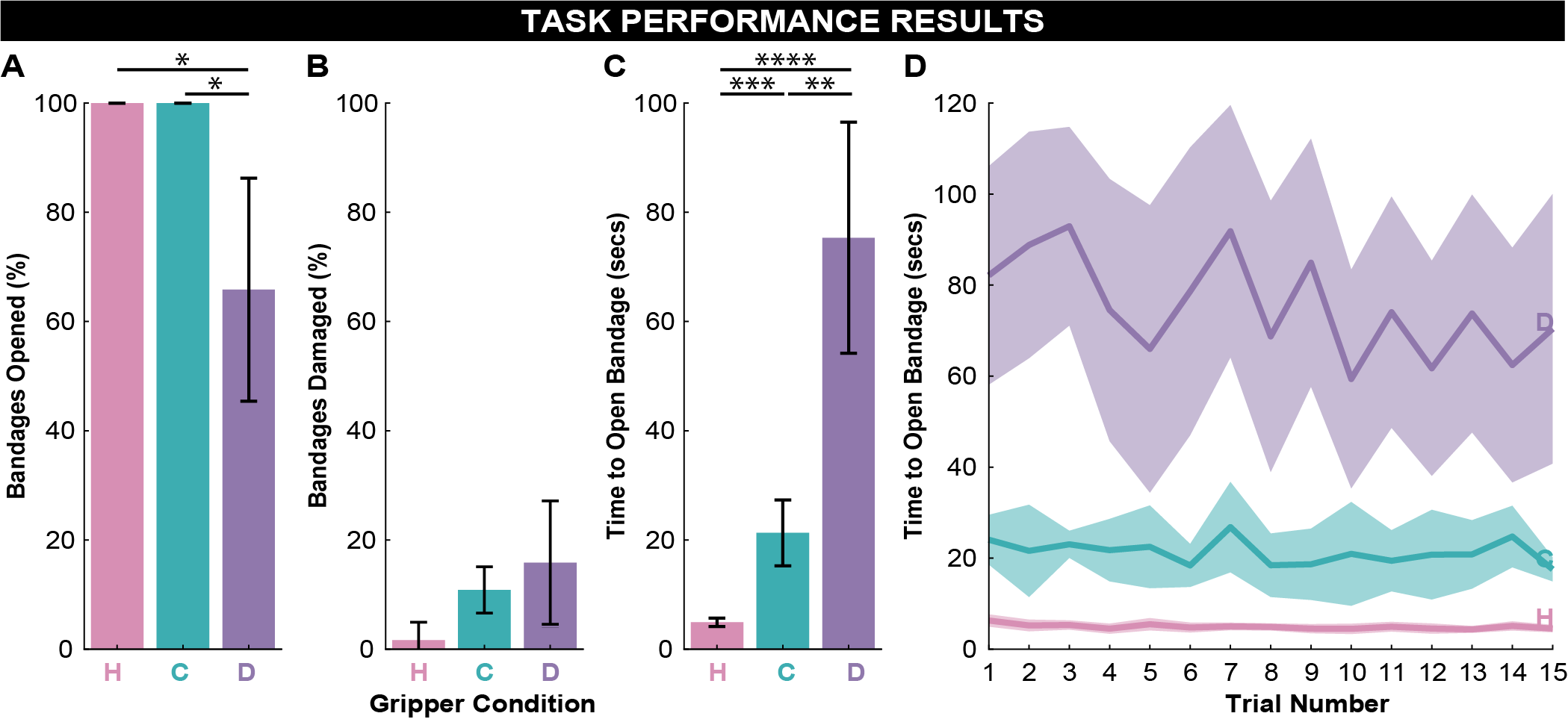}
   \caption{\textbf{Task Performance Results.} Bar graphs of task performance metrics (\textbf{A:} Bandages Opened, \textbf{B:} Bandages Damaged, and \textbf{C:} Time to Open Bandage) averaged across participants in each gripper condition (Hands, \textit{H}; Concentrated Load Grippers, \textit{C}; Distributed Load Grippers, \textit{D}). 
   \textbf{D:} Time to Open Bandage over trials averaged across participants in each gripper condition. Error bars and shaded regions represent the 95\% confidence intervals. *, **, ***, and **** indicate that the Bonferroni-corrected planned comparison between gripper conditions was statistically significant with $p\rm{_{adj}< 0.05, 0.01,0.005}$, and $\rm{0.001}$, respectively. }
   \label{fig:taskPerformance}
\end{figure*}

After these two initial design modifications, demonstrators were finally able to open a bandage using the UMI handheld grippers. This outcome underscored the critical role of mechanical design in demonstration performance and prompted us to investigate what additional design changes might further enhance performance.

To this end, we conducted a study, described in Sections \ref{section: methods} and \ref{section: results}, to examine how finger load paths influence demonstration quality. In UMI, contact forces are distributed along the length of the fingers, whereas in the ALOHA teleoperation system the fingers are angled six degrees inward, concentrating force at the fingertips (Figure \ref{fig: taskDescription}A). The ALOHA design has previously been shown to be effective for opening contact lens packages \cite{zhao2023learning}, suggesting that fingertip-focused force application may also improve performance in bandage opening.
\section{EXPERIMENTAL METHODS}
\label{section: methods}

An experiment was conducted to evaluate the usability of handheld gripper tools during a bandage opening task and assess how variations in handheld gripper mechanics affect the quality of demonstrations, measured by time and damage, as well as the demonstrator’s perceived workload.

\subsection{Participants} 

Eight participants (5 females, 3 males; age: 26.9 $\pm$ 5.2 years) were recruited for this study. Eligibility criteria included being between the ages of 18 and 65 years old, having no history of cognitive impairment, the ability to perform 30 minutes of light physical activity, and no prior experience working with the robotic technology used in the study. This study protocol was reviewed and approved by the Institutional Review Board of the University of Massachusetts Amherst (IRB \#3669).

\subsection{Experimental Design}

Participants performed 15 bimanual bandage package-opening trials with each of the following three gripper conditions (Figure \ref{fig: taskDescription}A): 
\begin{itemize}
    \item \textbf{Distributed Load Grippers (D)} – modified UMI grippers with parallel, stiff fingers that distribute contact force across the full finger surface when closed.  
    \item \textbf{Concentrated Load Grippers (C)} – modified UMI grippers with angled, stiff fingers that concentrate contact force at the fingertips when closed.  
    \item \textbf{Hands (H)} – participants opened bandages using their own hands without assistance.  
\end{itemize}

Note that since we are only interested in the performance of the human demonstrator, and because weight and physical discomfort were a concern, the cameras were not included on the grippers during testing. 

In each trial, the participant was instructed to pick up a wrapped bandage (1" $\times$ 3") from the ``start" location, unwrap the bandage as quickly as possible without damaging it, and place the bandage in the ``stop" location (Figure \ref{fig: taskDescription}B). Each trial lasted a maximum of two minutes. 

The order of the distributed load and concentrated load gripper conditions was counterbalanced across participants, and the hands condition was always performed last. After each gripper condition, participants were asked to complete the NASA Task Load Index (NASA-TLX) questionnaire to assess the subjective workload of the task. 

\begin{table*}[th]
\centering
\begin{tabular}{lll}
\hline
\textbf{Subscale Title} & \textbf{Description}                                                     & \textbf{ANOVA Results: Gripper Condition}\\ \hline
Mental Demand           & How mentally demanding was the task?                                     & $F(2, 12) = 22.33, p < 0.001$\\
Physical Demand         & How physically demanding was the task?                                   & $F(2, 12) = 13.71, p < 0.001$\\
Temporal Demand         & How hurried or rushed was the pace of task?                              & $F(2, 12) = 2.35, p = 0.14$\\
Performance             & How successful were you in accomplishing what you were asked to do?      & $F(2, 12) = 13.05, p < 0.001$\\
Effort                  & How hard did you have to work to accomplish your level of performance?   & $F(2, 12) = 21.01, p < 0.001$\\
Frustration             & How insecure, discouraged, irritated, stressed, and/or annoyed were you? & $F(2, 12) = 24.92, p < 0.001$\\ \hline
\end{tabular}
\caption{NASA-TLX Questions and ANOVA Results}
\label{table:nasatlxquestions}
\end{table*}

\subsection{Task Performance and Perceived Workload Metrics} 

Task performance was assessed across the three gripper conditions using the following metrics:  
\begin{itemize}
    \item \textbf{Bandages Opened} – the percentage of bandages successfully opened within the two-minute time limit.  
    \item \textbf{Bandages Damaged} – the percentage of bandages damaged during the task (e.g., bandage torn, backing removed).  
    \item \textbf{Time to Open Bandage} – the time required to open each bandage per trial (capped at two minutes).  
\end{itemize}

Perceived workload for each gripper condition was measured using the NASA-TLX questionnaire, a widely used and validated tool for measuring the overall workload associated with a given task \cite{hart1988development}. The NASA-TLX is comprised of six subscales, three of which relate to the demands imposed on the participant (Mental, Physical, Temporal) and three of which relate to the participant's interaction with the task (Effort, Frustration, Performance) (Table \ref{table:nasatlxquestions}). In this study, raw NASA-TLX scores were analyzed \cite{said2020validation}. Due to an administrative error, one participant did not complete the NASA-TLX questionnaire.

\subsection{Statistical Analyses} 

Within-subjects analysis of variance (ANOVA) was conducted to assess the effect of gripper condition -- and trial number, when appropriate -- on the task performance and perceived workload metrics. Statistically significant effects were followed up with planned comparisons in the form of two-tailed paired-samples $t$-tests with the Bonferroni adjustment (referred to as $p\rm{_{adj}}$). All statistical analyses were performed using a custom script in MATLAB (MathWorks, Natick, MA). For all statistical tests, the significance level was set to $\alpha=0.05$. 
\section{RESULTS}
\label{section: results}

\subsection{Bandages Opened} 
The one-way within-subjects ANOVA revealed a statistically significant effect of gripper condition on the percentage of bandages opened within the two-minute time limit $[F(2, 14) = 10.75, p = 0.002]$ (Figure \ref{fig:taskPerformance}A). Post hoc comparisons indicated that the percentage of bandages opened with the distributed load grippers was significantly lower than with hands $[t(7) = 3.28, p_{adj} = 0.041]$ and the concentrated load grippers $[t(7) = 3.28, p_{adj} = 0.041]$. Specifically, all bandages were successfully opened with hands and the concentrated load grippers, whereas only 65.8\% were successfully opened in the distributed load grippers.

\subsection{Bandages Damaged} 

The one-way within-subjects ANOVA found that the effect of gripper condition on the percentage of bandages damaged did not reach statistical significance $[F(2, 14) = 3.42, p = 0.062]$ (Figure \ref{fig:taskPerformance}B).

\subsection{Time to Open Bandage} 

The two-way within-subjects ANOVA revealed a statistically significant effect of gripper condition on the time to open bandage $[F(2, 14) = 32.28, p < 0.001]$ (Figure \ref{fig:taskPerformance}C). Post hoc comparisons revealed that bandages were opened significantly faster with hands compared to with concentrated load grippers $[t(7) = 5.39, p_{adj} = 0.003]$ and distributed load grippers $[t(7) = 6.98, p_{adj} < 0.001]$. Additionally, bandages were opened significantly faster with concentrated load grippers compared to the distributed load grippers $[t(7) = 4.56, p_{adj} = 0.008]$. 

Neither the effect of trial number, $[F(14, 98) = 1.36, p = 0.19]$, nor the interaction of gripper condition $\times$ trial number, $[F(28, 196) = 1.20, p = 0.24]$, reached statistical significance (Figure \ref{fig:taskPerformance}D).  

\subsection{Perceived Workload} 

\begin{figure*}
   \centering
   \vspace{1.5mm}
   \includegraphics[width =0.95 \textwidth]{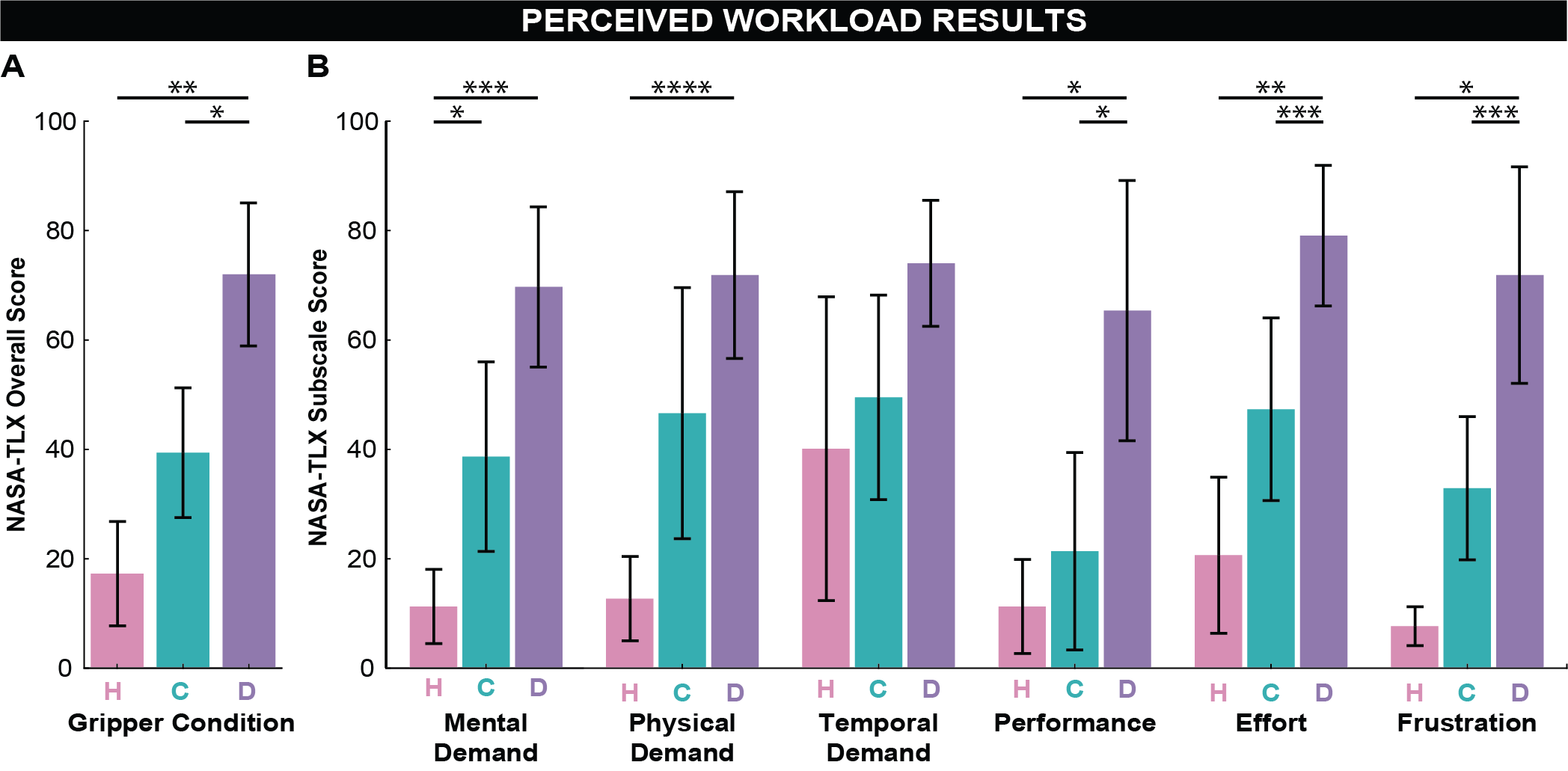}
   \caption{\textbf{Perceived Workload Results.} Bar graphs of task performance metrics (\textbf{A:} NASA-TLX overall raw score  and \textbf{B:} NASA-TLX subscale scores averaged across participants in each gripper condition (Hands, \textit{H}; Concentrated Load Grippers, \textit{C}; Distributed Load Grippers, \textit{D}). Error bars represent the 95\% confidence intervals. *, **, ***, and **** indicate that the Bonferroni-corrected planned comparison between gripper conditions was statistically significant with $p\rm{_{adj}< 0.05, 0.01,0.005}$, and $\rm{0.001}$, respectively.}
   \label{fig:perceivedWorkload}
\end{figure*}

The one-way within-subjects ANOVA revealed a statistically significant effect of gripper condition on the overall raw NASA-TLX score $[F(2, 12) = 19.6, p < 0.001]$ (Figure \ref{fig:perceivedWorkload}A). Post hoc comparisons indicated that perceived overall work load with the distributed load grippers was significantly higher than with hands $[t(6) = 5.39, p_{adj} = 0.005]$ and the concentrated load grippers $[t(6) = 4.41, p_{adj} = 0.013]$. 

One-way within-subjects ANOVAs on each of the NASA-TLX subscale scores revealed statistically significant effects of gripper condition on all subscale scores except Temporal Demand. The ANOVA results for each subscale are summarized in Table \ref{table:nasatlxquestions}.

Post hoc comparisons are summarized in Figure \ref{fig:perceivedWorkload}B. These results indicated that participants rated bandage opening with the distributed load grippers as significantly more mentally demanding than with either hands or the concentrated load grippers, and as more physically demanding compared to hands. No differences across conditions were observed for temporal demand, possibly due to the instructions the participants received (unwrap the bandage as quickly as possible) and the imposed 2-minute trial limit. Participants also rated performance as worse, and effort and frustration as higher, when using the distributed load grippers compared to hands or the concentrated load grippers.

\section{DISCUSSION}
\label{section: discussion}

Overall, the usability study demonstrated that altering the force distribution of UMI gripper fingers significantly affected participants’ ability to open bandage packages, with concentrated load grippers outperforming distributed load grippers. These findings highlight that subtle hardware changes can substantially improve demonstration quality and, in turn, the robot control policies learned from them. Accordingly, perceived workload was highest with the distributed load grippers. No significant differences were observed between hands and concentrated load grippers, except for mental demand, which was rated significantly lower when using hands. However, participants still opened bandages $15\times$ faster with their hands than with distributed load fingers and $4\times$ faster than with concentrated load fingers. Thus, even with performance gains, the 200 demonstrations required to train a UMI policy remain time-intensive.

Extensive research has focused on improving learning from demonstration through algorithmic approaches without altering mechanical design \cite{Barekatain_2024}. For example, Sakr et al. \cite{Sakr2025} showed that an augmented reality guidance system reduced the number of demonstrations required for a 2D maze navigation task, and Fu et al. \cite{fu2024mobilealohalearningbimanual} demonstrated that different imitation learning methods (e.g., ACT, Diffusion Policy, and VINN + Chunking) yield different performance outcomes using the Mobile ALOHA teleoperation system. However, the effectiveness of these algorithmic approaches remains constrained by a robot’s contact mechanics during interaction with the environment. As our results demonstrate, even a small change in finger load distribution can substantially affect baseline task performance. Thus, improving gripper mechanics is a fundamental component of robot learning from human demonstration that will only serve to complement and enhance the performance of existing learning algorithms and human demonstration methods.

We do, however, note that a key limitation of this work is that the UMI grippers used in the study were not equipped with the sensors or markers required for integration into a complete demonstration-to-robot-action pipeline. As a result, the findings cannot be directly generalized to the full learning from demonstration process. Instead, this study focused on evaluating the quality of demonstrations collected with different gripper designs and on exploring how mechanical modifications might reduce the performance gap between gripper-based demonstrations and those performed with bare hands. By isolating the effects of mechanical design, this work provides initial insights into how hardware refinements can enhance demonstration efficiency and user experience, which is a necessary step in addition to addressing sensing and control challenges in end-to-end robot learning. Future work will incorporate sensing and tracking capabilities into UMI to enable evaluation of the complete demonstration-to-robot pipeline in healthcare-relevant tasks.

Ultimately, our work reinforces the long-recognized principle that limitations in gripper hardware design cannot necessarily be overcome through improved control or learning algorithms alone\cite{hogan1985impedance}. Mechanical properties such as force distribution, stiffness, and ergonomics fundamentally shape what demonstrations are possible, and no amount of algorithmic refinement can compensate for hardware that is incapable of achieving the required interactions. In fact, handheld gripper tools may serve as a novel approach to disentangle whether poor robot grasping performance stems from deficiencies in the gripper design or shortcomings in the control policy. By allowing human demonstrators to perform tasks with grippers that closely mimic robotic end-effectors, researchers can isolate the impact of hardware limitations from that of learning algorithms. This dual role positions handheld tools not only as data collection devices but also as valuable diagnostic instruments for guiding both hardware and control co-design in robot learning.

For now, efficient bandage opening will require further design refinements to improve both human demonstration quality and subsequent robot performance. However, we emphasize that UMI as is still shows the potential to be a very valuable resource, and still may have a place as a tool to bridge healthcare and robotic technology, depending on the task. In fact, the open-source nature, including the exceptional build guide, the complete pipeline, and the robot-agnostic functionality, is what makes UMI an excellent candidate for exploring robotics in real-world, complicated settings such as healthcare. As new and improved handheld grippers emerge, we encourage continued evaluation of their usability, ergonomics, and impact on demonstration quality, both for targeted applications such as medical package opening and for broader manipulation capabilities.

\section{CONCLUSION}
\label{section: conclusion}
This study demonstrates that small but targeted changes to handheld gripper mechanics can substantially influence human demonstration quality in learning from demonstration systems. In healthcare contexts, where professional demonstrators have limited time and task complexity is high, optimizing demonstration tools for force distribution could reduce the number of required trials and improve policy learning efficiency. These findings extend beyond the specific bandage opening task, suggesting a broader design principle to better align the mechanical properties of handheld learning from demonstration tools with the physical affordances of the target task, which can help bridge the gap between human and robot performance. Ultimately, this work is important because improving the usability of handheld grippers can reduce the effort required of healthcare workers, enable more accurate demonstrations, and ultimately accelerate the development of robots capable of assisting in clinical practice.


%%%%%%%%%%%%%%%%%%%%%%%%%%%%%%%%%%%%%%%%%%%%%%%%%%%%%%%%%%%%%%%%%%%%%%%%%%%%%%%%

%%%%%%%%%%%%%%%%%%%%%%%%%%%%%%%%%%%%%%%%%%%%%%%%%%%%%%%%%%%%%%%%%%%%%%%%%%%%%%%%

%%%%%%%%%%%%%%%%%%%%%%%%%%%%%%%%%%%%%%%%%%%%%%%%%%%%%%%%%%%%%%%%%%%%%%%%%%%%%%%%

\bibliographystyle{IEEEtran}
\bibliography{ICRA25bib}

\end{document}